\documentclass[11pt,a4paper]{article}
\usepackage[hyperref]{acl2020}
\usepackage{times}
\usepackage{latexsym}

\usepackage{makecell}
\usepackage{multirow}
\usepackage{amsmath}
\usepackage[misc]{ifsym} 

\usepackage{microtype}

\usepackage{graphicx}
\graphicspath{ {./appendix/} }

\aclfinalcopy 

\title{Distilling Knowledge from Pre-trained Language Models via Text Smoothing}

\author{Xing Wu\footnotemark[1] , Yibing Liu\footnotemark[1] , Xiangyang Zhou$^{(\textrm{\Letter})}$ \and Dianhai Yu \\Baidu Inc., Beijing, China \\ \{wuxing03, liuyibing01, zhouxiangyang, yudianhai\}@baidu.com}

\date{}

\begin{document}
\maketitle
\footnotetext{Equally contributed.}
\begin{abstract}

This paper studies compressing pre-trained language models, like BERT~\citep{devlin-etal-2019-bert}, via teacher-student knowledge distillation.
Previous works usually force the student model to strictly mimic the smoothed labels predicted by the teacher BERT.
As an alternative, we propose a new method for BERT distillation, i.e., asking the teacher to generate smoothed word ids, rather than labels, for teaching the student model in knowledge distillation.
We call this kind of method \textbf{Text Smoothing}.
Practically, we use the softmax prediction of the Masked Language Model (MLM) in BERT to generate word distributions for given texts and smooth those input texts using that predicted soft word ids.
We assume that both the smoothed labels and the smoothed texts can implicitly augment the input corpus, while text smoothing is intuitively more efficient since it can generate more instances in one neural network forward step.
Experimental results on GLUE and SQuAD demonstrate that our solution can achieve competitive results compared with existing BERT distillation methods.

\end{abstract}

\section{Introduction}

\begin{figure}[h]
\centering
\includegraphics[width=0.49\textwidth,height=0.19\textheight]{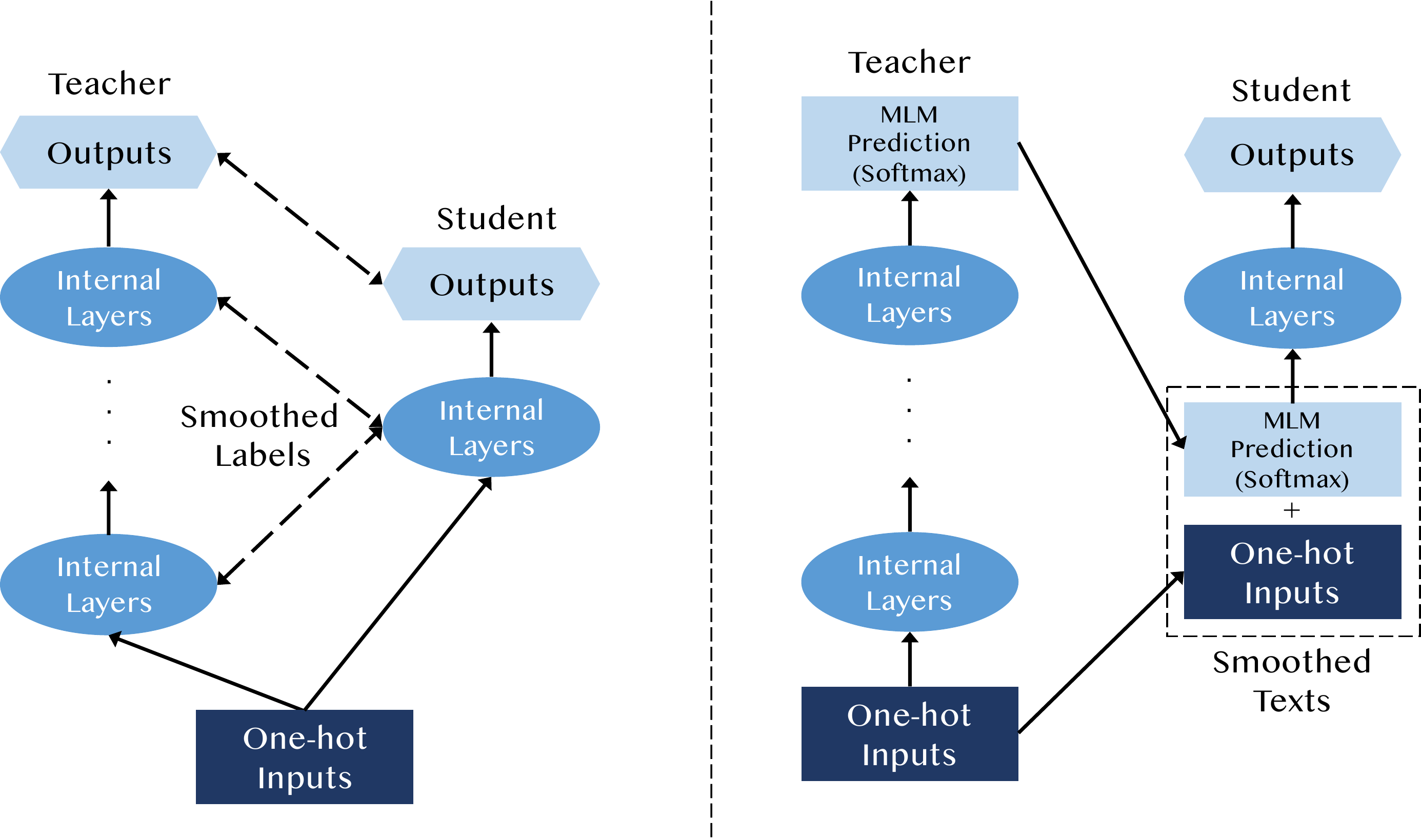}
\caption{Distilling BERT using smoothed labels (left) and smoothed texts (right).}
\label{fig:1}
\end{figure}

While language model pre-training, such as BERT \citep{devlin-etal-2019-bert} and its variants \citep{NIPS2019_8812,liu2019roberta,lan2019albert,raffel2019exploring}, has significantly improved the performance of many natural language processing tasks, those pre-trained models are usually too large to be deployed for resource-limited applications.
To address this problem, many researchers recently investigate using the Knowledge Distillation (KD) algorithms~\citep{hinton2015distilling} to transfer the knowledge of a large pre-trained language model (teacher) into a small neural network (student) \citep{sanh2019distilbert,sun-etal-2019-patient,jiao2019tinybert}, in order to reduce the model size for online deployment.
The left part in Figure ~\ref{fig:1} shows the learning algorithm of previous KD methods for BERT.
Typically, the student model learns from smoothed labels, i.e., given an input text of some downstream task, both BERT and the student model are asked to conduct prediction over the input text, the student model is then asked to strictly fit the soft labels predicted by BERT (for example, the probability distribution over different labels for text classification tasks). 

As an alternative, we investigate training a student model using smoothed texts, rather than labels, in knowledge distillation of BERT.
The right part of Figure ~\ref{fig:1} demonstrates our proposed method, which is comprised of two steps:
\begin{itemize}
\item \textbf{Text Smoothing}: Given the pre-trained teacher (BERT) as well as the downstream task corpus, we first feed the one-hot encoding of each text in the corpus into the teacher BERT, and fetch the softmax prediction of the Masked Language Model (MLM) in BERT, the original one-hot encodings and the predicted word distributions are then mixed up together to smooth the raw texts.  
\item \textbf{Student Learning}: The student model is tuned using the smoothed corpus of the downstream task, akin to typical supervised training. During the inference stage, the student model only takes the one-hot encoding as input to make predictions.
\end{itemize}
We assume that both the label smoothing and text smoothing can implicitly generate more instances for training the student model, while text smoothing could be more efficient since it can intuitively generate more instances than label smoothing in each forward step of neural networks (because the number of word candidates is much larger than task labels).
Our work is inspired by the recent success of using soft word ids to improve performance in Neural Machine Translation~\citet{NIPS2019_8861,gao-etal-2019-soft}, and we extend this idea to knowledge distillation of BERT.

Empirical results on GLUE~\citep{DBLP:journals/corr/abs-1804-07461} and SQuAD~\citep{DBLP:journals/corr/RajpurkarZLL16,rajpurkar-etal-2018-know} show that the student model that trained with our proposed text smoothing algorithm can achieve competitive results, compared with existing BERT KD methods, while the text smoothing method is significantly faster than previous distillation algorithms.
The major contribution of our work is that we propose a new knowledge distillation method for BERT, which uses BERT to generate smoothed texts, rather than labels, for teaching the student model.

\section{Related Work}

Compressing large pre-trained models like BERT attracts increasing research interests in recent years.
Existing studies can be roughly divided into \emph{distillation}~\citep{sanh2019distilbert,sun-etal-2019-patient,jiao2019tinybert}, \emph{pruning}~\citep{mccarley2019pruning} and \emph{quantization}~\citep{zafrir2019q8bert}, where the distillation method shows the most promising results.
In this paper, we focus on the knowledge distillation of BERT.
Most previous BERT distillation methods are based on making the student model mimic the smoothed labels predicted by the teacher BERT.
\citet{sanh2019distilbert} use the large-scale corpus in pre-training to generate a massive amount of weak labels to teach the student model, to improve the performance of the student model.
\citet{sun-etal-2019-patient} investigate forcing the student model fitting multiple feedforward layers in the teacher BERT to improve the performance of the student model.
\citet{jiao2019tinybert} propose a more systemic approach to construct an equally-effective student of BERT, where they use the feedforward, multi-head attention distribution and embedding layers of the teacher model to supervise the student model,  conduct the KD method on both pre-training and fine-tuning sides, and leverage Data Augmentation (DA)~\citep{wu2019conditional} algorithms to increase size of fine-tuning corpus in 20 times.

\section{Our Method}

Let $\mathcal{T}$ be the teacher BERT and $\mathcal{S}$ be the student model to be tuned.
Given the dataset of some downstream task, namely $\mathcal{D} = \{t_i, p_i, s_i, l_i\}_{i=1}^N$, where $N$ is the number of instances, $t_i$ is the one-hot encoding of a text (a single sentence or a sentence pair), $p_i$ is the positional encoding of $t_i$, $s_i$ is the segment encoding of $t_i$ and $l_i$ is the label of this instance.
We train the student model of BERT in two steps:

\textbf{Text Smoothing}: We feed the one-hot encoding $t_i$, positional encoding $p_i$ as well as the segment encoding $s_i$ into BERT, and fetch the output of the last layer of the transformer encoder in BERT, which is denoted as:
\begin{align}
\overrightarrow{t_i} = \text{BERT}(t_i) \\
\overrightarrow{t_i} \in \mathcal{R}^{\text{seq\_len}, \text{emb\_size}}
\label{bert-encoder}
\end{align}
\noindent
where $\overrightarrow{t_i}$ is a 2D dense vector in shape of [\text{sequence\_len}, \text{embedding\_size}].
We then multiply $\overrightarrow{t_i}$ with the word embedding matrix $W \in \mathcal{R}^{\text{vocab\_size,} \text{embed\_size}}$ in BERT, to get the MLM prediction results, which is defined as:
\begin{align}
\text{MLM}(t_i) = \text{softmax}(\overrightarrow{t_i}W^T)
\label{mlm-softmax}
\end{align}
\noindent
where each row in $\text{MLM}(t_i)$ is a probability distribution over the word vocabulary, representing the choice of words in that position of the input text learned by BERT.
It is worth noticing that we do not apply the mask corruption in BERT in fetching the MLM prediction in text smoothing, besides the teacher BERT is also not fine-tuned with the downstream task corpus.
Our experimental results show that the mask corruption may harm the performance in knowledge distillation.
The input text $t_i$ is finally smoothed using a simple linear interpolation as:
\begin{align}
\widetilde{t_i} = \lambda \cdot t_i + (1-\lambda) \cdot \text{MLM}(t_i).
\label{text-smoothing}
\end{align}
\noindent
where $\lambda$ controls the smoothing degree.

\begin{table*}
\centering
\small
\begin{tabular}{ |c|c|c|c|c|c|c|c|c|c|c|c| } 
 \hline
Method (\# layers) &  MNLI-m  & MNLP-mm  & QQP & SST-2 & QNLI & MRPC & RTE & CoLA & STS-B & Average \\ \hline
 \hline
BERT$_{base}$ (12) & 84.6 & 83.4 & 71.2 & 93.5 & 90.5 & 88.9 & 66.4 & 52.1 & 85.8 & 79.6 \\
\hline
BERT$_{small}$ (3) & 74.8 & 74.3 & 65.8 & 86.4 & 84.3 & 80.5 & 55.2 & 16.8 & 67.5 & 68.3 \\
BERT-PKD (3) & 76.7 & 76.3 & 68.1 & 87.5 & 84.7 & 80.7 & 58.2 & - & - & - \\
TextSmooth (3) & 77.2 & 76.1 & 66.7 & 88.4 & 85.2 & 82.3 & 60.7 & 23.8 & 77.7 & 70.9 \\
\hline
BERT$_{small}$ (4) & 75.4 & 74.9 & 66.5 & 87.6 & 84.8 & 83.2 & 62.6 & 19.5 & 77.1 & 70.2 \\
DistillBERT (4) & 78.9 & 78.0 & 68.5 & 91.4 & 85.2 & 82.4 & 54.1 & 32.8 & 76.1 & 71.9\\
BERT-PKD (4) & 79.9 & 79.3 & 70.2 & 89.4 & 85.1 & 82.6 & 62.3 & 24.8 & 79.8 & 72.6\\
TinyBERT (4) & 82.5 & 81.8 & 71.3 & 92.6 & 87.7 & 86.4 & 62.9 & 43.3 & 79.9 & 76.5\\
TinyBERT w/o DA (4) & 80.5 & 81.0 & - & - & - & 82.4 & - & 29.8 & - & -\\
TextSmooth (4) & 79.9 & 79.2 & 69.6 & 90.6 & 86.1 & 85.0 & 63.3 & 33.3 & 79.8 & 74.1\\
\hline
BERT$_{small}$ (6) & 80.4 & 79.7 & 69.2 & 90.7 & 86.7 & 85.9 & 63.6 & 30.6 & 81.9 & 74.3 \\
BERT-PKD (6) & 81.5 & 81.0 & 70.7 & 92.0 & 89.0 & 85.0 & 65.5 & - & - & -\\
TinyBERT (6) & 84.6 & 83.2 & 71.6 & 93.1 & 90.4 & 87.3 & 70.0 & 51.1 & 83.7 & 79.4\\
TinyBERT w/o DA (6) & - & - & - & - & - & - & - & - & - & -\\
TextSmooth (6) & 81.9 & 80.9 & 70.3 & 92.8 & 88.0 & 86.4 & 65.7 & 42.7 & 82.8 & 76.8\\
\hline
\end{tabular}
\caption{Experimental results on the GLUE benchmark test set. TextSmooth is our method.}
\label{table:results-glue}
\end{table*}

\begin{table}
\centering
\small
\begin{tabular}{|c|cccc|}
    \hline
    \multicolumn{1}{|c|}{Method (\# layers)} &
    \multicolumn{2}{c}{SQuAD 1.1} &
    \multicolumn{2}{c|}{SQuAD 2.0} \\
    \multicolumn{1}{|c|}{\quad} &
    \multicolumn{1}{c}{EM} &
    \multicolumn{1}{c}{F1} &
    \multicolumn{1}{c}{EM} &
    \multicolumn{1}{c|}{F1} \\
    \hline
    BERT$_{base}$ (12) & 80.7 & 88.4 & 73.1 & 76.4 \\
   \hline
   BERT$_{small}$ (4) & 67.8 & 77.5 & 60.0 & 63.9 \\
   BERT-PKD (4) & 70.1 & 79.5 & 60.8 & 64.6 \\
   DistillBERT (4) & 71.8 & 81.2 & 60.6 & 64.1 \\
   TinyBERT (4) & 72.7 & 82.1 & 65.3 & 68.8 \\
   TextSmooth (4) & 70.9 & 80.5 & 62.9 & 66.2 \\
   \hline
   BERT$_{small}$ (6) & 76.5 & 84.7 & 65.7 & 69.0 \\
   BERT-PKD (6) & 77.1 & 85.3 & 66.3 & 69.8 \\
   DistillBERT (6) & 78.1 & 86.2 & 66.0 & 69.5 \\
   TinyBERT (6) & 79.7 & 87.5 & 69.9 & 73.4 \\
   TextSmooth (6) & 76.7 & 84.8 & 66.7 & 69.5 \\
  \hline
\end{tabular}
\caption{Experimental results on the SQuAD dev set. The result of TinyBERT w/o DA is not reported, so we skip those missing scores.}
\label{table:results-squad}
\end{table}

\textbf{Student Learning}: Given those smoothed texts, we construct the corpus for teaching the student model, denoted as $\mathcal{\widetilde{D}} = \{\widetilde{t_i}, p_i, s_i, l_i\}_{i=1}^N$. The student model is then tuned with $\mathcal{\widetilde{D}}$ as supervised learning.
In the inference phase, the student model takes the one-hot encoding of text as input and conducts predictions, the same as most existing teacher-student KD methods.

\section{Experiments}

\subsection{Dataset and Metrics}
Following the previous works, we test the performance of our proposed KD method on two large-scale datasets: the GLUE benchmark~\citep{DBLP:journals/corr/abs-1804-07461} and SQuAD (1.1 and 2.0)~\citep{DBLP:journals/corr/RajpurkarZLL16,rajpurkar-etal-2018-know}, covering eleven different tasks in natural language processing. The evaluation metrics in this paper are also the same as previous works.

\subsection{Baselines}
Same as the previous works, we take the pre-trained BERT$_{base}$ (12-layers, English, uncased, no whole word masking) as our teacher model, and we use a smaller BERT (contains fewer layers of transformer encoder) as the student model to be tuned.
The baselines to be compared include DistillBERT~\citep{sanh2019distilbert}, BERT-PKD~\citep{sun-etal-2019-patient} and TinyBERT~\citep{jiao2019tinybert}, those kinds of methods are all based on mimicking soft labels, while TinyBERT is more complicated and uses Data Augmentation (DA) to boost the performance of the student model.
To make a fair comparison, we also take TinyBERT w/o DA as another baseline, to better analyze the difference between label smoothing and text smoothing in BERT distillation.
Additionally, we study directly fine-tuning the student model without any smoothing (denoted as BERT$_{small}$), as ablation experiments, to validate the performance of all knowledge distillation algorithms.

\subsection{Experiment Setting}
We test 3 different student models: 3-layer BERT, 4-layer BERT, and 6-layer BERT, following the previous works.
All the other hyper-parameters are the same as the teacher BERT and the $\lambda$ is set to 0.5 in our experiments.
Akin to the previous works, the student model is initialized by reusing the weights of first n layers of the transformer encoder as well as the embedding layers in the teacher BERT. 
We directly copy the experimental results of our baselines if reported, otherwise, we use `-' to represent those missing scores or directly skip that setting if the baseline methods do not consider this setting in their experiments.
\begin{figure*}[h]
\centering
\includegraphics[width=0.65\textwidth,height=0.111\textheight]{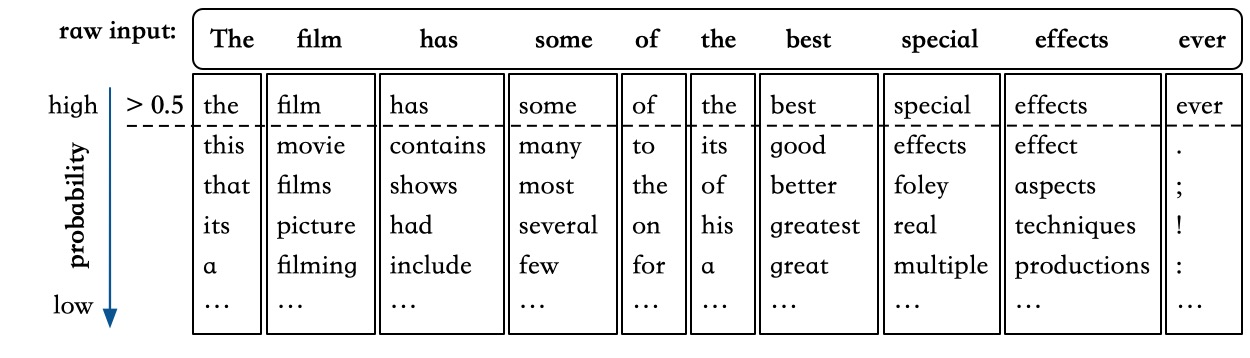}
\caption{Five unique fake sentences generated for the raw input text ``\emph{The film has some of the best special effects ever}" using text smoothing.}
\label{fig:2}
\end{figure*}

\subsection{Experimental Results}
Table ~\ref{table:results-glue} and Table ~\ref{table:results-squad} show the experimental results of our proposed distillation method as well as all baseline and ablation methods.
As demonstrated, the text smoothing method can consistently over-perform existing smooth label based KD methods in different student models and most natural language processing tasks, including DistillBERT, BERT-PKD as well as TinyBERT w/o DA, especially for small student models.
The performance of TinyBERT is significantly improved when using data augmentation, which also implies the importance of smoothing the texts, instead of the labels. 
The performance of the ablation model BERT$_{small}$ is significantly worse than both the label smoothing based methods as well as our text smoothing based methods, which implies the effectiveness of knowledge distillation algorithms in BERT. Besides the performance, our proposed text smoothing distillation method is also faster than the other methods.
We only use about 92 seconds to pass one epoch in GLUE on average, while BERT-PKD needs to use more than 233 seconds on the same devices and deep learning framework.
Moreover, our proposed method can also support transferring the teacher knowledge to heterogeneous students, since we do not rely on mimicking the intermediate layers of the teacher BERT, which increases the searching space for the optimal student model.

\section{Analysis}
To better understand how text smoothing works and where is the limitation, we visualize the smoothing representation of an example in the SST-2 corpus by sampling some fake sentences from its smoothed text representation. 
Figure ~\ref{fig:2} shows the visualization results, the top sentence (``\emph{The film has some of the best special effects ever}") is a raw sentence in SST-2, and the five sentences below are all generated by randomly sampling words at each position of the raw input sentence, using its smoothed word distribution.
The probability of generating the whole sentence is calculated using the multiplication of the probability for generating each word.

As we can observe, all those sampled sentences are semantically-related to the original input text, the higher the generation probability is, the more semantically similar to the raw text the generated sentence is.
Those sampled sentences are surprisingly diverse in expression (such as \emph{has-contains}, \emph{effects-foley}, \emph{some-several} in Figure ~\ref{fig:2}).
An interesting thing is that the top one generated sentence is usually the raw input itself, as shown in Figure ~\ref{fig:2}.
The reason may be the MLM in BERT pre-training, which facilitates the reconstruction and paraphrasing of the input sentence.
However, many of those generated sentences have some small errors in grammar (like ``\emph{the of}" of the third sentence) and some of them suffer from semantic shifts (like ``\emph{productions}" of the last sentence). 
Those kinds of errors may harm the performance of some tasks who needs to consider fine-grained semantic information or be aware of syntactic structures of the input sentences.
Studying the impact of those errors and how to eliminate them shall be an interesting research topic in the future.

\section{Conclusion}
In this paper, we propose a new knowledge distillation method for compressing BERT.
Instead of making the student model mimic the soft labels predicted by the teacher BERT, we explore leveraging the Masked Language Model (MLM) in BERT to generate word distributions for each text in the downstream task corpus, and using those predicted word distributions to smooth the raw input text.
The student model then learns from those smoothed texts, rather than smoothed labels.
Experimental results on GLUE and SQuAD show that our text smoothing method can achieve competitive results compared with most existing BERT KD methods.
In the future, we would like to take a deep look at the principle of text smoothing and try to combine text smoothing and label smoothing for better knowledge distillation of language model pre-training.

\bibliography{acl2020}
\bibliographystyle{acl_natbib}

\end{document}